\pdfoutput=1

\documentclass[11pt]{article}

\usepackage[]{template/acl}

\usepackage{times}
\usepackage{latexsym}

\usepackage[T1]{fontenc}

\usepackage[utf8]{inputenc}

\usepackage{microtype}
\usepackage{graphicx}
\usepackage{tabularx}
\usepackage{booktabs}
\usepackage{caption}
\usepackage{subcaption}

\newcommand{\mpara}[1]{\medskip\noindent{\bf #1}}

\title{The Effect of Masking Strategies on Knowledge Retention by Language Models}

\author{Jonas Wallat$^\clubsuit$ \and Tianyi Zhang$^\clubsuit$ \and Avishek Anand$^{\clubsuit,\spadesuit}$\thanks{This work was done while Avishek was at L3S.} \\
        $^\clubsuit$L3S Research Center Hannover, Germany \\ $^\spadesuit$TU Delft, The Netherlands \\ \texttt{<firstname.lastname>@l3s.de}}

\begin{document}
\maketitle
\begin{abstract}
Language models retain a significant amount of world knowledge from their pre-training stage. This allows knowledgeable models to be applied to knowledge-intensive tasks prevalent in information retrieval, such as ranking or question answering. Understanding how and which factual information is acquired by our models is necessary to build responsible models. 
However, limited work has been done to understand the effect of pre-training tasks on the amount of knowledge captured and forgotten by language models during pre-training.
Building a better understanding of knowledge acquisition is the goal of this paper. 
Therefore, we utilize a selection of pre-training tasks to infuse knowledge into our model. In the following steps, we test the model's knowledge retention by measuring its ability to answer factual questions.
Our experiments show that masking entities and principled masking of correlated spans based on pointwise mutual information lead to more factual knowledge being retained than masking random tokens. Our findings demonstrate that, like the ability to perform a task, the (factual) knowledge acquired from being trained on that task is forgotten when a model is trained to perform another task (catastrophic forgetting) and how to prevent this phenomenon. To foster reproducibility, the code\footnote{https://github.com/jwallat/knowledge-acquisition}, as well as the data\footnote{https://github.com/facebookresearch/PAQ} used in this paper, are openly available.
\end{abstract}


\section{introduction}

Overparameterized language models (LMs) such as BERT~\citep{DBLP:conf/naacl/DevlinCLT19} or T5~\citep{DBLP:journals/jmlr/RaffelSRLNMZLL20} have resulted in breakthroughs in a wide variety of language tasks.
One of the main reasons for the success of such models is the large parametric memory that results in the memorization of factual knowledge during pre-training~\citep{DBLP:conf/naacl/PetroniPFLYCTJK21}. 
This large parametric memory is often used to better transfer performance on many knowledge-intensive tasks such as question answering, fact-checking, and knowledge-base completion.

Consequently, there have been recent studies that aim to measure the amount of factual knowledge in the parametric memory of such large language models by \textit{probing} \citep{DBLP:journals/corr/abs-2102-12452}. 
Probing entails training small classifiers on the LM's representations to predict linguistic properties such as part-of-speech tags \citep{DBLP:conf/acl/TenneyDP19} or factual knowledge \citep{DBLP:conf/emnlp/PetroniRRLBWM19}. 
However, a key demerit of probing is that, although it can establish the presence of knowledge, it does not guarantee that the knowledge \textit{can be} or \textit{is} indeed used successfully in downstream tasks such as question answering \citep{DBLP:journals/tacl/BelinkovG19, DBLP:conf/emnlp/TamkinSGG20}.
In this paper, we are interested in measuring the presence of such actionable factual knowledge.

Our approach to investigating knowledge acquisition and containment is to use \textit{closed-book question answering} (CBQA) as a representative task.
The aim of CBQA is to generate answers to input questions just by using the parametric memory of the language model under consideration.
The distinct advantage of using CBQA as a task, instead of \textit{probing}, is that we are operating purely on the LM parameters and not relying on external output layer parameters.
Second, CBQA is a well-defined downstream task and hence a reasonably good proxy for measuring actionable factual knowledge contained in the parametric memory of LMs. 

\begin{figure*}[ht]
    \centering
    \includegraphics[width=0.98\textwidth]{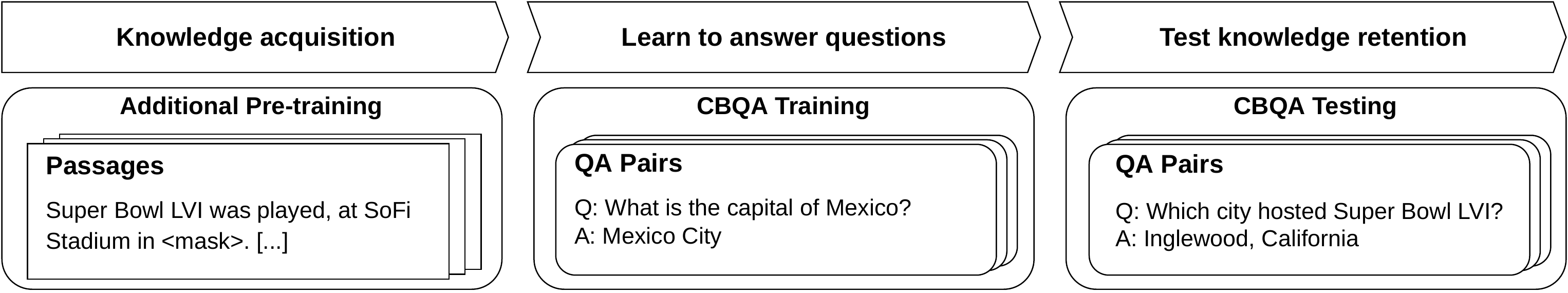}
    \caption{Experimental setup. First, the model is additionally pre-trained on knowledge-intense passages from Wikipedia. It is trained on closed-book question answering and finally tested on QA pairs directly built from the pre-training passages to measure the amount of knowledge retained.}
    \label{fig:experimental_setup}
\end{figure*}

Knowledge is acquired by LMs by carrying out large-scale pre-training tasks or pre-fine-tuning tasks before the actual fine-tuning over the downstream task.
Recent work by \citet{DBLP:conf/emnlp/RobertsRS20} and \citet{DBLP:journals/corr/LewisPAQ} used an additional round of pre-training to infuse more knowledge into CBQA models.
However, it has been well known that fine-tuning over downstream tasks results in catastrophic forgetting. 
The second objective of this paper is to systematically study how factual knowledge is forgotten under different training regimes.
We examine knowledge containment and forgetting under multiple \textit{knowledge-infusion tasks} - \textit{random token masking}  \citep{DBLP:journals/jmlr/RaffelSRLNMZLL20}, \textit{salient span masking}~\citep{DBLP:conf/acl/ZhangHLJSL19}, and \textit{PMI masking}~ \citep{DBLP:conf/iclr/LevineLLALTS21}. 
Specifically, we use the PAQ dataset \citep{DBLP:journals/corr/LewisPAQ}, which contains \textit{Wikipedia passages} and associated question-answer pairs for our experiments. 
The Wikipedia passages are used as data for our \textit{knowledge-infusion tasks}, and the associated QA pairs are used to compose the CBQA dataset.
This direct relation between a pre-training corpus and the CBQA dataset allows for a systematic investigation of different pre-training tasks and their effect on the amount of factual knowledge retained. 
Our approach can be succinctly summarized as a three-step process (see Figure~\ref{fig:experimental_setup}): First, knowledge is infused into a LM (here T5) using the PAQ Wikipedia passages followed by an optional CBQA fine-tuning.
Then, the CBQA held-out sets evaluate the degree of knowledge contained and retained.

\subsection{Summary of Contributions}

In this paper, we attempt to answer the previously less understood question of how actionable factual knowledge is acquired and forgotten under different knowledge-infusion strategies.
Specifically, we answer these two research questions:


\mpara{\textbf{RQ I.}} What is the effect of pre-training tasks on the amount of knowledge being acquired?

\mpara{\textbf{RQ II.}} Is this knowledge still usable after fine-tuning for downstream tasks?
\\

We conduct an extensive study on the utilized Wikipedia passages and corresponding QA pairs, first training the model on the passages and later testing the amount of knowledge captured via the QA pairs (see Figure~\ref{fig:experimental_setup}). Our key findings are summarized as follows:

\mpara{Pre-training tasks.}  
Our results suggest that SSM (+9.4\%) and PMI masking (+8.9\%) both outperform random token masking (+4.4\%), with PMI masking having the least amount of variance between runs. 



\mpara{Mitigation of catastrophic forgetting.} 
We find that applying regularizers such as elastic weight consolidation (EWC) \citep{DBLP:journals/corr/KirkpatrickPRVD16} and multi-task learning (MTL) can mitigate catastrophic forgetting of factual knowledge. 
MTL seems slightly more effective than EWC, yet both result in substantial performance improvement over the baseline (26.8\% and 23.4\%). Additionally, we find that MTL and EWC reduce the run performance variance by 88\% and 75.8\%, respectively.

\section{Related Work}

In the past, language models have been shown to acquire an understanding of linguistic features and knowledge from being trained on natural language corpi (an overview can be found in the survey by \citet{DBLP:journals/tacl/BelinkovG19}). Interestingly, these models also remember factual information from their training stage. \citet{DBLP:conf/emnlp/PetroniRRLBWM19} studied the amount of relational knowledge in BERT and found it to be similar to common knowledge-base completion models. Subsequent works are built on this finding and have tried different methods to localize factual knowledge. \citet{DBLP:journals/corr/abs-2104-08696} used influence functions on neuron activations to find subsets of neurons responsible for predictions about specific entities. \citet{wallat-etal-2020-bertnesia} analyzed all of BERT's layers, finding that most of the factual knowledge resides in later layers - the same ones changed the most by fine-tuning the model. Other recent work analyzing the architecture of transformer-based language models found that feed-forward layers act as key-value stores, storing shallow features in the lower layers and more sophisticated semantics in the upper layers \citep{DBLP:conf/emnlp/GevaSBL21}. 

By retaining knowledge from training, language models can be applied to a set of knowledge-intensive tasks such as knowledge graph completion (e.g., \citep{DBLP:journals/corr/abs-1909-03193}) and increasingly to closed-book question answering (CBQA) \citep{DBLP:conf/emnlp/RobertsRS20, DBLP:journals/corr/LewisPAQ}. Other than reading comprehension (e.g., SQuAD \citep{DBLP:conf/emnlp/RajpurkarZLL16}) or Open-QA (e.g., NaturalQuestions \citep{DBLP:journals/tacl/KwiatkowskiPRCP19}), CBQA requires the model to produce the answer without the help of an outside context, directly from the model's parametric memory \citep{DBLP:conf/nips/LewisPPPKGKLYR020}. 
Recently, CBQA models have been further improved by exposing them to an additional pre-training phase (\citep{DBLP:conf/emnlp/RobertsRS20,  DBLP:journals/corr/LewisPAQ} inter alia). While \citet{DBLP:conf/emnlp/RobertsRS20} showed how parametric memory scales with model size, \citet{DBLP:journals/corr/LewisPAQ} were able to achieve similar results with significantly smaller models, by training on knowledge-dense text. In contrast to the related work, our goal is not achieving SotA performance on CBQA benchmarks but studying how we can optimize the additional pre-training step to infuse more knowledge into language models. No systematic study could be identified that investigates pre-training tasks and the amount of knowledge retained and forgotten.

Much of the previous work defaults to masking entities (e.g., \citep{DBLP:conf/emnlp/RobertsRS20,  DBLP:journals/corr/LewisPAQ}), given the success of entity masking in REALM \citep{DBLP:journals/corr/GuuREALM}. In their work, \citet{DBLP:journals/corr/GuuREALM} augment the pre-training of their Open-QA model with a knowledge retriever, providing factual context to answer questions. They find entity masking \citep{DBLP:conf/acl/ZhangHLJSL19} to outperform masking random tokens \citep{DBLP:conf/naacl/DevlinCLT19} or spans \citep{DBLP:journals/tacl/JoshiCLWZL20} for their Open-QA knowledge retriever model. 
Since the REALM model learns to retrieve and not necessarily store knowledge in its parameters, it is unclear if the effectiveness of entity masking translates into pre-training for CBQA. 
Closely related, \citet{DBLP:conf/emnlp/YeLWBMYRK21} investigate the effectiveness of additional pre-training with different masking strategies to improve downstream task performance on a variety of tasks - finding that the usage of direct supervision or meta-learning to learn masking strategies can sometimes outperform heuristic approaches.  
We also aim to shed light on the effectiveness of masking strategies, especially on the amount of knowledge that is retained and forgotten from pre-training.

\section{Knowledge Infusion Tasks}
To infuse knowledge and test retention, we follow the methodology depicted in Figure~\ref{fig:experimental_setup}.
We start with (1) pre-training on the passages with one of our three pre-training tasks (Section~\ref{sec:infusing_knowledge}), followed by (2) CBQA fine-tuning on the training set of corresponding QA pairs. 
Lastly, (3) we measure the amount of knowledge retained from pre-training by the difference to a baseline model that was not additionally pre-trained. 
All experiments are repeated five times. To keep our experiments tractable, we use T5-small with 12 layers.

\subsection{Additional Pre-training}
\label{sec:infusing_knowledge}
To infuse knowledge into the model, we continue pre-training T5 on the \textit{language modeling} objective.  
We use the cross-entropy loss as our \textit{knowledge-infusion loss} (cf. Equation~\ref{formula:loss}) with a batch size of $32$ and a maximum number of $100$ epochs (with early-stopping). 
We experiment with the following masking strategies:

\mpara{Random token masking (RTM).} In the RTM task, we apply the standard T5 language modeling task of span corruption. We randomly select positions and mask $15\%$ tokens or spans, as detailed in the original T5 paper \citep{DBLP:journals/jmlr/RaffelSRLNMZLL20}. 

\mpara{Salient span masking (SSM).} Other than random spans, SSM prioritizes masking named entities \citep{DBLP:conf/acl/ZhangHLJSL19}. As the answers in QA are mostly related to entities (e.g., "Who is the current French president?"), SSM could have a positive inductive bias by putting additional focus on entities. In our experiments, we mask 15\% of the tokens. If there are fewer entities in the sequence, we mask all available entities and default back to masking tokens randomly until we reach 15\% masked tokens. We use a BERT-based named entity recognizer \citep{DBLP:journals/corr/GuuREALM, DBLP:conf/emnlp/RobertsRS20}.

\mpara{Pointwise-Mutual Information (PMI) masking.} The last of the masking strategies, PMI masking, was proposed by~\citet{DBLP:conf/iclr/LevineLLALTS21}. The authors find that LMs can perform well in the RTM objective by focusing on local features rather than understanding the general context.
Therefore, PMI masking jointly masks frequently co-located n-grams in the training corpus \citep{DBLP:conf/iclr/LevineLLALTS21}. PMI masking has two benefits for acquiring knowledge: (1) The likelihood of masking non-informative tokens decreases (tokenizing ``by the way'' as one instead of three tokens), and (2) the context of entities has to be understood when masking ``the united states'' as one token since it will not allow an easy guess of ``the united [MASK]''.


\subsection{CBQA Fine-tuning}
Before measuring how much knowledge is retained by the model, we fine-tune T5 for CBQA, allowing the model to learn how to answer questions. Therefore, we split our QA pairs into a training set and a holdout test set. T5 models all tasks as sequence-to-sequence problems. Thus, similar to the knowledge infusion loss, the fine-tuning also uses the cross-entropy: 
\begin{equation}
    \mathcal{L}_{QA}= \mathcal{L}_{KI} =-\sum_{i=1}^{|V|}y_{i}log(p_i)
    \label{formula:loss}
\end{equation}
with the vocabulary size $|V|$ and $p_i$ being the softmax probability of the individual tokens. For the CBQA fine-tuning, we use a batch size of 128 and train for a maximum of 100 epochs (with early stopping). At inference time, we generate answers using beam search decoding with five beams.  
After fine-tuning, we use the hold-out test set of QA pairs to test how much knowledge is retained. We measure the CBQA performance of the different knowledge-infused models and a baseline without additional pre-training and report exact match scores. If a model, after being pre-trained on the passages, improves performance over the baseline, we say it acquired and retained knowledge from the pre-training process.

\subsection{Data}
We use the PAQ dataset \citep{DBLP:journals/corr/LewisPAQ} for our experiments. This dataset contains a set of 65M QA pairs automatically generated from Wikipedia passages, with each passage having 4-8 QA pairs. 
Since the QA pairs are derived from a passage, we can be sure that the answer to the question can be found in that passage. Another benefit is that our QA pairs cover multiple facts in one passage, allowing for more precise measurements of what is retained. To keep our experiments tractable, we select 10K passages from the full PAQ dataset. For the additional pre-training, we utilize the full 10K passages. For the CBQA training, we split the ~49k QA pairs into the train (76\%), dev (10\%), and test sets (14\%). An example of a PAQ passage with corresponding QA pairs can be found in Table~\ref{tab:paq_example}.


\begin{table*}
    \centering
    \begin{tabularx}{\linewidth}{lc}
        \toprule
        \textbf{Passage} \\
        \midrule
        \multicolumn{2}{l}{\textit{The Witcher, by Polish writer Andrzej Sapkowski, is a fantasy series of short stories and novels about}} \\
        \multicolumn{2}{l}{\textit{the witcher Geralt of Rivia. In Sapkowski's books, "witchers" are monster hunters who (with training  }} \\
        \multicolumn{2}{l}{\textit{and body modification) develop "supernatural abilities at a young age to battle deadly beasts. [...]}} \\
        \midrule
        \textbf{Question} & \textbf{Answer} \\
        \midrule
        Who is the author of the witcher series? & Andrzej Sapkowski \\
        Who is the author of the witcher series? & Polish \\
        What kind of books are the witcher series? & fantasy \\
        What are the roles of witchers in the book? & monster hunters \\
        \bottomrule
    \end{tabularx}
    \caption{Example passage and corresponding QA pairs from the PAQ dataset. We utilize the fact that 4-8 QA pairs cover each passage for a more precise measurement of the acquired knowledge.}
    \label{tab:paq_example}
\end{table*}
\subsection{Effect of Pre-training Tasks}


The first research question we address is the effect of pre-training tasks on the knowledge acquired by the language model. Thus, we trained a baseline, which was only fine-tuned, and three different models that were first pre-trained on their respective tasks (RTM, SSM, PMI) and then fine-tuned. The results in exact match and F1 score are reported in Table~\ref{tab:initial_results_tasks}. 

\begin{table}[ht]
\centering
\begin{tabular}{lccc} \hline
             & \textbf{EM}     & \textbf{F1}     & \textbf{Gain} \\ \hline
FT  & $51.22_{\pm0.55}$ & $57.28_{\pm0.64}$ & -                    \\ \hline
RTM & $53.47_{\pm0.82}$ & $59.28_{\pm0.79}$ & 4.4\%                 \\
SSM & $56.05_{\pm1.71}$ & $61.53_{\pm1.68}$ & 9.4\%                 \\
PMI & $55.78_{\pm0.62}$ & $61.65_{\pm0.52}$ & 8.9\%                 \\ \hline
\end{tabular}
\caption{CBQA performance for the baseline without additional pre-training (i.e., only fine-tuning, FT) and three models with additional pre-training and different masking strategies (RTM, SSM, PMI).}
\label{tab:initial_results_tasks}
\end{table}

First, we find the process of additional pre-training to improve the amount of knowledge retained for the downstream task, regardless of the specific masking strategy. 
This is in line with other works, reporting a positive effect of additional pre-training on the CBQA performance \citep{DBLP:journals/corr/GuuREALM, DBLP:conf/emnlp/RobertsRS20, DBLP:journals/corr/LewisPAQ, DBLP:conf/emnlp/YeLWBMYRK21}. 
Furthermore, we observe that PMI masking (+8.9\%) and SSM (+9.4\%) outperform masking random tokens (+4.4\%). 
We hypothesize that this is the case because RTM results in many easy-to-guess examples being masked (as discussed by \citet{DBLP:conf/iclr/LevineLLALTS21} and in Section~\ref{sec:infusing_knowledge}). By masking easy-to-guess examples, the model is never required to understand the broader context in which entities appear. 
Facts - or relational knowledge - are often expressed in the interactions and context of entities. 
Since the answers to a large portion of the questions (e.g., "Who won Super Bowl 50?") are mostly entities, it is reasonable that SSM outperforms RTM. 
PMI masking finds co-occurring tokens and masks them together, making it likely that multiple tokens building a surface form of an entity will also be masked together, resulting in a similar performance to SSM. 
Furthermore, while the average performance of SSM and PMI is fairly similar, we observe less variance between runs with PMI masking compared to SSM (-87\%) or RTM (-43\%). 
Given this increase in training stability with similar performance, PMI masking might be a suitable alternative for infusing knowledge into language models.

\paragraph{\textbf{Insight 1.}} PMI and SSM-based losses retain more knowledge than random masking.

\subsection{Knowledge Forgetting}

Next, we address our second research question: is the acquired knowledge usable for downstream tasks?
Our results presented in Table~\ref{tab:initial_results_tasks} suggest that at least part of the knowledge can be used in CBQA. Yet, given the extensive pre-training on passages containing the facts, the improvement is relatively small. 
Thus, we wonder if the sequential learning problem \citep{McCloskey1989CatastrophicII} (also referred to as catastrophic forgetting) applies to the factual knowledge stored in the model's parametric memory. 
The classical sequential learning problem occurs when a model is trained on an initial task, fine-tuned on another, and required to perform the initial task again. In this scenario, many works have shown that the model is likely to (partially) lose its ability to perform the initial task \citep{McCloskey1989CatastrophicII, DBLP:conf/iclr/MosbachAK21, DBLP:journals/corr/KirkpatrickPRVD16, DBLP:conf/coling/GuF20}. We hypothesize that this applies to the (factual) knowledge acquired with the initial task. 
In a parallel effort, \citet{DBLP:journals/corr/abs-2110-03215} investigated knowledge forgetting of language models when exposed to new training data (although not changing training objectives), finding that continual learning techniques such as regularization and parameter expansion are useful to balance the dropping of existing knowledge over new information. 
Therefore, we experimented with common approaches to alleviate the sequential learning problem. Still, other than \citet{DBLP:journals/corr/abs-2110-03215}, we investigate whether catastrophic forgetting occurs when training objectives are changed. 

\subsection{Elastic Weight Consolidation.}
In the first set of experiments, we extend our training setup by using the elastic weight consolidation regularizer \citep{DBLP:journals/corr/KirkpatrickPRVD16} in the fine-tuning stage. As EWC tries to preserve weights that are important for the initial task, we hypothesize that it might also preserve the world knowledge acquired with this task. After including the EWC regularization term, our adapted loss function is

\begin{equation}
    \mathcal{L}_{EWC}(\theta)=\mathcal{L}_{QA}(\theta)+\sum_{i}{\frac{\lambda}{2}F_i(\theta_i-\theta_{KI,i}^{*})^2},
\end{equation}
where $\mathcal{L}_{QA}$ is the CBQA loss, $\lambda$ the regularization strength, $F$ the diagonal of the (empirical) Fisher Information Matrix (FIM), and $\theta_{KI,i}^{*}$ the model parameters after the knowledge infusion stage. 
EWC uses the approximated FIM to weigh how important the parameters are for the first (i.e., knowledge infusion) task. It is defined by:

\begin{equation}
    F = \frac{1}{N} \sum_{i=1}^{N}{\nabla_{\theta}log \: p(x_i|\theta)\nabla_{\theta}log \: p(x_i|\theta)^{T}}
\end{equation}

where $x_i$ are sampled training instances \citep{DBLP:journals/jmlr/Martens20}. We experimentally select a regularization strength of 1000 and keep the rest of our setup as described before. 
For a more detailed explanation of how EWC works, we refer to the original paper \citep{DBLP:journals/corr/KirkpatrickPRVD16} and the more theoretical description by \citet{DBLP:journals/corr/abs-2105-04093}.



\begin{figure*}[ht]
    \centering
    \begin{subfigure}[b]{0.49\textwidth}
         \centering
         \includegraphics[width=\textwidth]{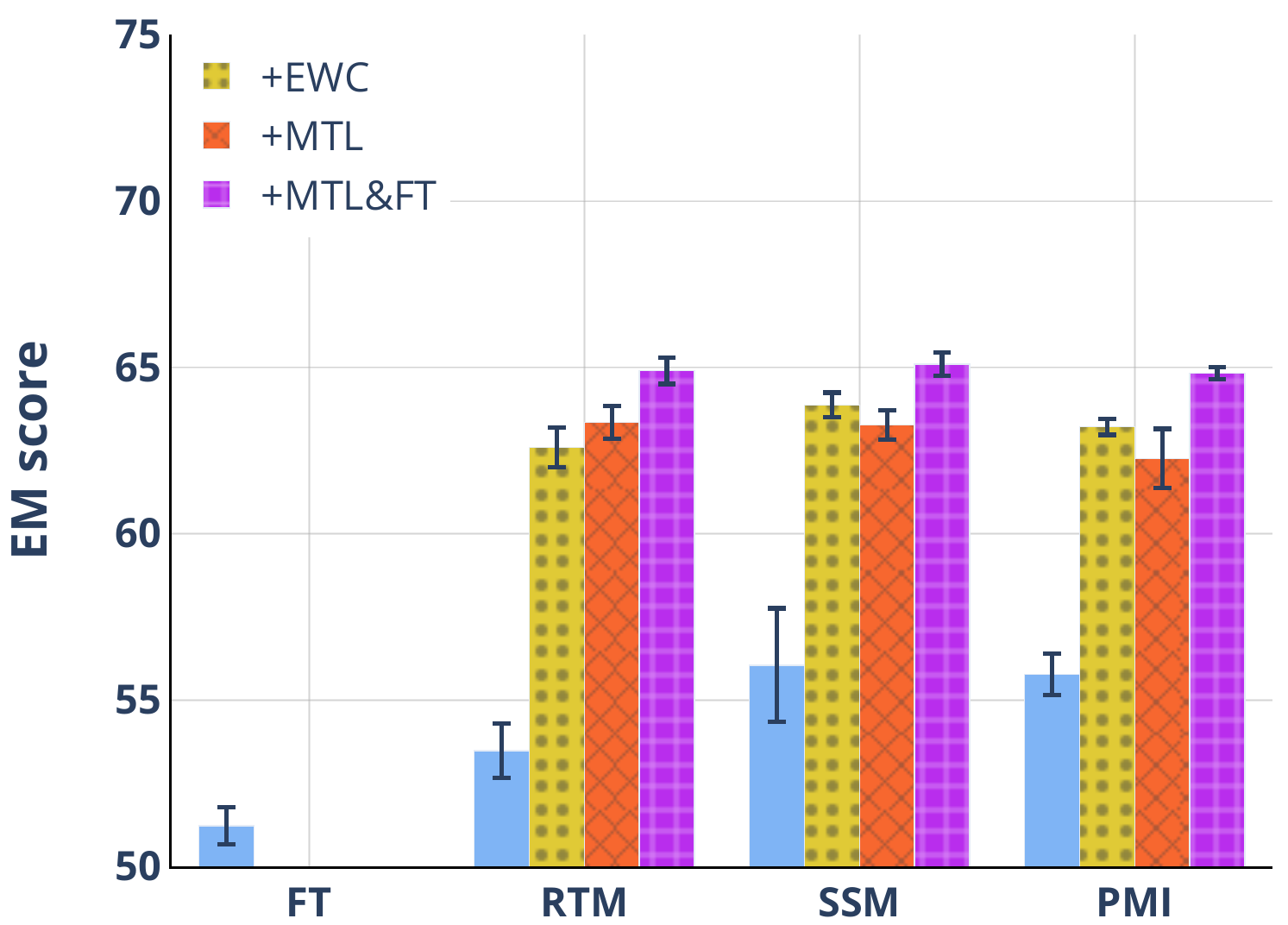}
         \caption{EM}
    \end{subfigure}
    \hfill
    \begin{subfigure}[b]{0.49\textwidth}
         \centering
         \includegraphics[width=\textwidth]{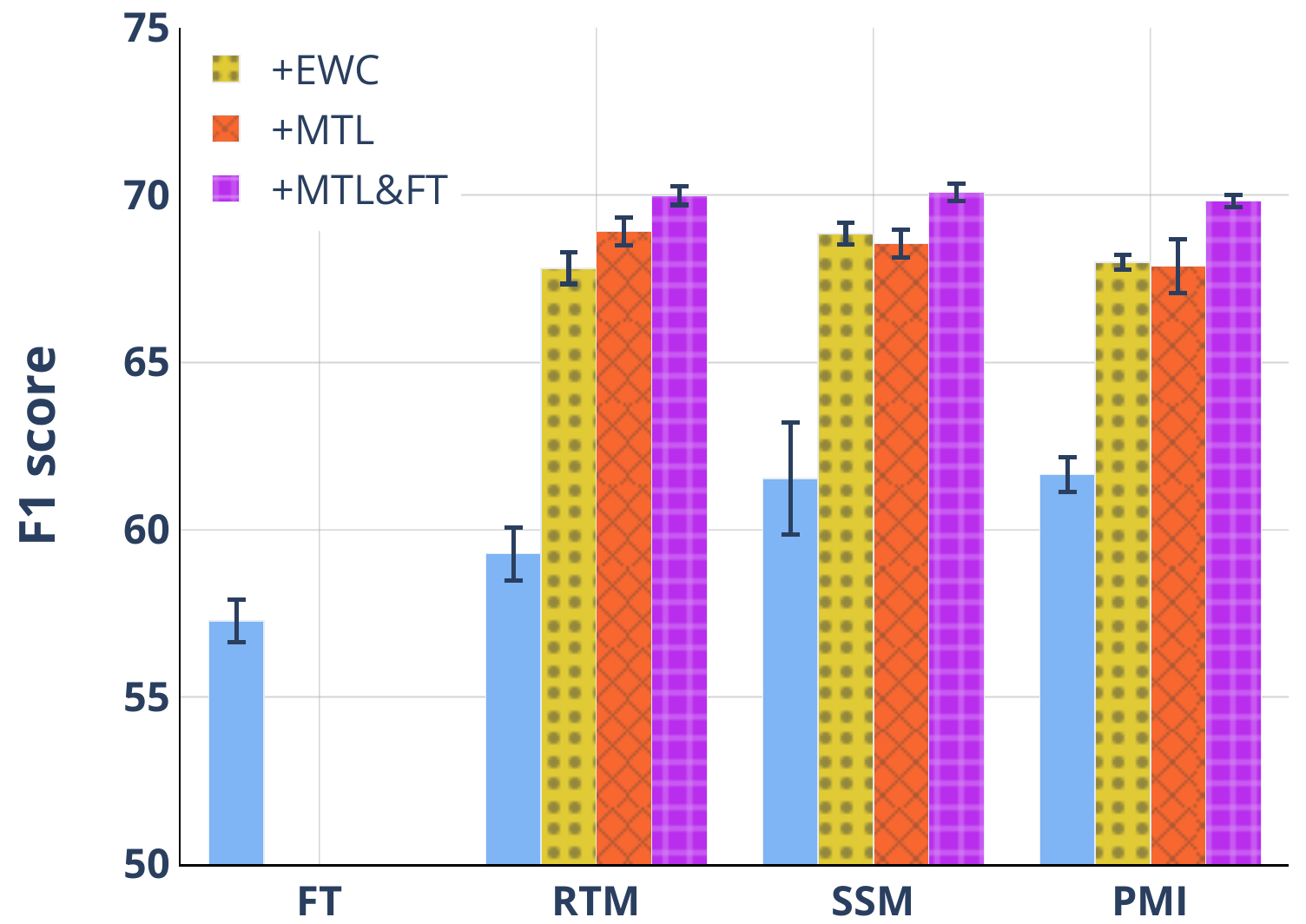}
         \caption{F1}
    \end{subfigure}
    \caption{CBQA performance for the baseline (only fine-tuning, FT) and three models with additional pre-training on different masking strategies + fine-tuning (RTM, SSM, PMI). Additionally, we show the effect of the elastic weight consolidation regularizer (+EWC), multi-task learning the PT objective and CBQA in parallel (+MTL) and MTL followed by another round of fine-tuning (+MTL\&FT).}
    \label{fig:results_big}
\end{figure*}

Figure~\ref{fig:results_big} compares the EM score achieved by the pre-training tasks with and without EWC. 
It is directly evident that the amount of knowledge retained and usable from the pre-training stage increases with the use of EWC (23.4\% on average). 
This suggests that the problem of catastrophic forgetting indeed applies to the knowledge captured from an initial task. 
In these experiments with EWC, we again find SSM and PMI masking to outperform RTM. 
However, the differences between masking strategies are not as striking as in the standard setting. 
Additionally, a benefit of EWC is the reduced amount of variance between runs with the same masking strategy.  
The substantial increase in performance suggests that a large part of the acquired knowledge is indeed forgotten and, in the unregularized setting, not usable for CBQA.


\subsection{Multi-task Learning.}
Due to the success of EWC in counteracting the forgetting of factual knowledge, we next experiment with multi-task learning.
MTL has been shown to work well in retaining the ability to perform multiple tasks (\citep{DBLP:conf/gcai/RibeiroMD19} inter alia). 
By not sequentially training the model on two tasks, MTL forces the model to keep the ability to perform well in both tasks simultaneously. 
Thus, we compare our standard setup with two different training strategies: Multi-task training, both the additional pre-training and the CBQA fine-tuning together, as well as the aforementioned multi-task setup, followed by an additional CBQA fine-tuning step. We alternate batches between tasks and keep our setup as described before. 
The results are shown in Figure~\ref{fig:results_big} (MTL and MTL\&FT).



Similarly to EWC, we find that MTL outperforms the sequential training approach. The standard MTL setup performs similarly to EWC; however, these results can be further improved by another round of fine-tuning (+3.1\%). With this additional fine-tuning, we also observe the least amount of variance between runs. The results emphasize that in unregularized sequential training, a substantial part of the factual knowledge is forgotten. 

\paragraph{\textbf{Insight 2.}} EWC and MTL regularizers increase the amount of factual knowledge usable for downstream tasks. 
\section{Discussion and Conclusion}



In this paper, we study the effect of pre-training tasks on the amount of knowledge retained in the model. To do so, we use the PAQ dataset, which contains a large set of QA pairs and passages that contain ground-truth facts. 
In our experiments, we make several observations: First, in the current typical approach of additional pre-training followed by CBQA fine-tuning salient span masking outperforms random token masking. 
On average, the principled approach of PMI masking performs similarly to entity masking, with less variance between runs. 
Next, we hypothesize that sequentially training on language modeling followed by CBQA leads to factual knowledge no longer being accessible.
Therefore, we apply elastic weight consolidation and multi-task learning, two common strategies to counter catastrophic forgetting. 
We find catastrophic forgetting of knowledge indeed to be a problem, as both applying EWC and MTL leads to substantial improvements in actionable knowledge retained. 
We hope this study will lead to more research on knowledge acquisition and forgetting in language models.

\section*{Acknowledgements}
This research was funded by the Federal Ministry of Education and Research (BMBF), Germany under the project LeibnizKILabor with grant No. 01DD20003.

\bibliography{main}
\bibliographystyle{template/acl_natbib}




\end{document}